\documentclass{article}

\usepackage{microtype}
\usepackage{graphicx}
\usepackage{subcaption}
\usepackage{booktabs}

\usepackage{hyperref}

\usepackage[preprint]{icml2026}

\usepackage{amsmath}
\usepackage{amssymb}
\usepackage{mathtools}
\usepackage{amsthm}

\usepackage[capitalize,noabbrev]{cleveref}

\usepackage{bm}
\usepackage{algorithm}
\usepackage{algorithmic}
\usepackage[table]{xcolor}
\definecolor{Ecol}{RGB}{113,166,81}
\definecolor{Pcol}{RGB}{121,181,230}
\definecolor{Vcol}{RGB}{213,204,105}
\definecolor{Ccol}{RGB}{60,96,123}

\theoremstyle{plain}

\theoremstyle{definition}

\theoremstyle{remark}

\usepackage[textsize=tiny]{todonotes}

\usepackage[markup=default, arthormarkup=None]{changes}
\definechangesauthor[name={wuxiuping}, color=green]{r1}
\definechangesauthor[name={yuzhao}, color=orange]{r2}
\definechangesauthor[name={chengyuxin}, color=blue]{r3}

\icmltitlerunning{InjectRBP: Steering Large Language Model Reasoning Behavior via Pattern Injection
}

\begin{document}

\twocolumn[
  \icmltitle{InjectRBP: Steering Large Language Model Reasoning Behavior via Pattern Injection}

  \icmlsetsymbol{equal}{*}

  \begin{icmlauthorlist}
    \icmlauthor{Xiuping Wu}{equal,soton}
    \icmlauthor{Zhao Yu}{equal,xjtu}
    \icmlauthor{Yuxin Cheng}{equal,hku}
    \icmlauthor{Ngai Wong}{hku}
    \icmlauthor{Liangjun Ke}{xjtu}
    \icmlauthor{Tapas Mishra}{soton}
    \icmlauthor{Konstantinos V.Katsikopoulos}{soton}

  \end{icmlauthorlist}

  \icmlaffiliation{xjtu}{Xi'an Jiaotong University}
  \icmlaffiliation{soton}{University of Southampton}
  \icmlaffiliation{hku}{University of Hongkong}

  \icmlcorrespondingauthor{Konstantinos V. Katsikopoulos}{K.Katsikopoulos@soton.ac.uk}
  \icmlcorrespondingauthor{Tapas Mishra}{T.K.Mishra@soton.ac.uk}

  \icmlkeywords{Machine Learning, ICML}

  \vskip 0.3in
]

\printAffiliationsAndNotice{\icmlEqualContribution}

\begin{abstract}
Reasoning can significantly enhance the performance of Large Language Models. While recent studies have exploited behavior-related prompts adjustment to enhance reasoning, these designs remain largely intuitive and lack a systematic analysis of the underlying behavioral patterns. Motivated by this, we investigate how models' reasoning behaviors shape reasoning from the perspective of behavioral patterns. We observe that models exhibit adaptive distributions of reasoning behaviors when responding to specific types of questions, and that structurally injecting these patterns can substantially influence the quality of the models' reasoning processes and outcomes. Building on these findings, we propose two optimization methods that require no parameter updates: InjectCorrect and InjectRLOpt. InjectCorrect guides the model by imitating behavioral patterns derived from its own past correct answers. InjectRLOpt learns a value function from historical behavior-pattern data and, via our proposed \textit{Reliability-Aware Softmax Policy}, generates behavioral injectant during inference to steer the reasoning process. Our experiments demonstrate that both methods can improve model performance across various reasoning tasks without requiring any modifications to model parameters, achieving gains of up to 5.34\% and 8.67\%, respectively.
\end{abstract}

\section{Introduction}
Large language models (LLMs) have demonstrated strong performance on reasoning tasks such as mathematical problem-solving,  knowledge Q\&A, and coding \cite{deepseekai2024deepseekv3technicalreport,guo_deepseek-r1_2025,qwen3technicalreport,openai2025gptoss120bgptoss20bmodel,rein2024gpqa,aime25,lightman2023lets,austin2021program}. 
Through prompting and post-training techniques, the model can perform extensive intermediate reasoning before generating the final response. These approaches further enhance the performance of LLMs on reasoning tasks \cite{zero-shot-reasoner-2022,cot-elicit-2022,tot-2023,guo_deepseek-r1_2025}.

Reasoning behaviors play a crucial role in eliciting the LLM’s reasoning, regulating the reasoning process, and ultimately improving the quality of the final answers.
The Chain-of-thought (CoT) reasoning process is elicited by inserting phrases such as “Let’s think step by step” and “Wait” into the prompt, which, in turn, trigger the model’s reasoning and thereby lead to higher-quality final responses \cite{zero-shot-reasoner-2022}.
The “aha” moments observed during DeepSeek-R1 training provide evidence that LLMs can learn to use reasoning behaviors to actively shape their reasoning processes, thereby improving response quality through reinforcement learning \cite{guo_deepseek-r1_2025}.
Due to the autoregressive nature of LLMs, the reasoning behavior could affect the subsequent reasoning content. 
By appropriately modulating the model's reasoning behavior, one can intervene in its reasoning process.
Moreover, \citet{Yang2025TesttimePI} presented a framework that alleviates redundant reasoning by dynamically guiding and controlling the model’s reasoning behaviours. Furthermore,
they proposed a method that monitors model behavior and dynamically terminates the reasoning chain to reduce the cost of reasoning \cite{yang2025dynamic}. To improve the reasoning performance of LLMs, 
\citet{wang-etal-2025-prejudge} introduces a prejudge node into the reasoning behaviours.
These studies influence the model’s reasoning process and its final responses by manipulating the model’s behavior during reasoning.
However, these studies have predominantly proposed intuitive methods, lacking a systematic analysis of model reasoning behaviors.

This paper systematically investigates how models’ reasoning behaviors shape their final responses from the behavior pattern perspective. We first extract the reasoning behavior chain from the reasoning process. Then we conduct a comparative statistical analysis of the behavior patterns across multiple models and benchmarks, revealing that models exhibit adaptive reasoning behaviors when responding to specific types of questions. 
Next, by cross-injecting these behavior patterns across various tasks, we empirically show that injected patterns can substantially influence the quality of models’ reasoning processes and outcomes.
Building on these findings, we design two behavior pattern improvement methods, InjectCorrect and InjectRLOpt, that deliberately steer models toward more effective reasoning patterns, improving the model performance across various reasoning tasks without requiring any parameter modifications.
The main contribution of this paper is as follows:

\begin{itemize}
    \item We analyze the patterns of reasoning behavior of different models from the perspective of behavior pattern, revealing that LLM models exhibit adaptive reasoning behaviors when answering different types of questions. 
    \item We cross-inject behavioral patterns across datasets and observe that such an injection could substantially affect the final response quality.
    \item Building on the above findings, we design two methods, InjectCorrect and InjectRLOpt, to improve the model’s reasoning behaviors, which achieve gains of up to 5.34\% and 8.67\% accuracy improvement without parameter modifications, respectively.
\end{itemize}

\section{Related Work}

\textbf{Reasoning Process Analysis.}
Recent studies, such as OpenAI's o1 \cite{openai2025gptoss120bgptoss20bmodel} and DeepSeek-R1 \cite{guo_deepseek-r1_2025}, have introduced a new paradigm where models are incentivized to engage in extended thinking time through reinforcement learning. These models exhibit emergent behavioral patterns, such as the Aha moment—a spontaneous re-evaluation and correction of previous steps \cite{guo_deepseek-r1_2025}. Other research on Tree of Thoughts \cite{tot-2023} and Quiet-STaR \cite{zelikman2024quietstarlanguagemodelsteach} explores how searching and backtracking serve as structured behaviors for navigating complex logical spaces.

Beyond simple prompting, researchers found that the quality of these steps is paramount. \citet{lightman2023lets} demonstrated that process supervision—rewarding correct intermediate steps rather than just the final answer—leads to more robust reasoning and higher accuracy in mathematical domains. Inspired by this, \citet{golovneva_roscoe_2022} and \citet{prasad-etal-2023-receval} propose to evaluate reasoning quality at the level of individual steps, focusing on both correctness and informational content. They show that such fine-grained evaluation can effectively improve performance on downstream tasks. 

Our work examines the reasoning process through the lens of behavioral patterns, focusing on systematic patterns of reasoning behavior and their relationship to reasoning quality.

\textbf{Reasoning Behavior Improvement.} \citet{ICLR2024_84b686f7} demonstrated that LLMs extend beyond pure reasoning to accurately simulate human decision-making behaviors. Building on this, recent works enhance reasoning performance by explicitly reshaping these behavioral patterns.
\citet{pang2025reasoningpatternmatterslearning} improved capabilities by training models to mimic human reasoning patterns, while \citet{wang-etal-2025-prejudge} introduces a prejudge strategy, adding a dedicated prejudge node before full reasoning and fine-tuning the model to perform this prejudgment step. This approach leads to notable gains in reasoning quality by encouraging the model to first assess the problem before engaging in detailed reasoning. Conversely, to mitigate inefficient behaviors, \citet{Yang2025TesttimePI} introduced a test-time prompt intervention framework that dynamically steers the reasoning process to mitigate redundant reasoning chains, substantially shortening chains of thought while preserving performance. \citet{muennighoff-etal-2025-s1} showed that forcing extended thinking processes ("budget forcing") improves outcomes, whereas \citet{yang2025dynamic} proposed a dynamic termination method that stops generation when the model becomes highly confident in a candidate answer, thereby adaptively controlling reasoning depth. 

These methods modify the model’s reasoning behavior to influence the final responses. However, they primarily treat behavior patterns as levers for control rather than as objects of systematic study. In contrast, our work explicitly analyzes the model’s reasoning behavior patterns, examines how these patterns affect models, and designs methods to improve models' reasoning behaviors.

\begin{figure*}
    \centering
    \includegraphics[width=\textwidth]{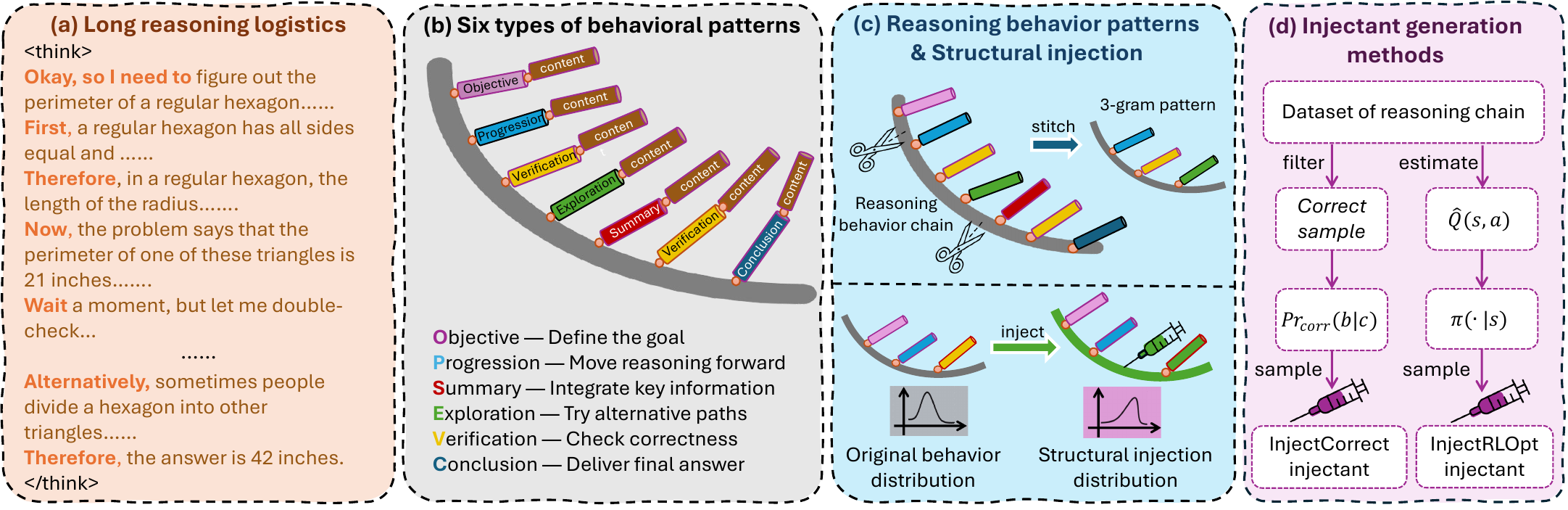}
    \caption{The analysis process of reasoning behavior pattern, including how the reasoning processes are decomposed from reasoning text into a structured “Reasoning Behavior Chain” and “Reasoning DNA”, and how frequent n‑gram behavior patterns are clipped and structurally injected into models to enhance their systematic reasoning ability.}
    \label{fig:reasoning_dna}
\end{figure*}

\section{Preliminaries}

\subsection{Reasoning Language Model}

A language model can be parameterized by $\theta$ and predicts the probability of a sequence $\mathbf{x}=(x_1,..,x_t)$. The joint probability is decomposed into a product of conditional next-token probabilities:
\begin{equation} \label{eq:def lm}
P_{\theta}(\mathbf{x}) = \prod_{t=1}^{T} P_{\theta}(x_t \mid \mathbf{x}_{<t}).
\end{equation}

In this work, we focus on the reasoning process, where for a given query $Q$, the model is required to generate an intermediate reasoning trajectory $R$ before producing the final answer $A$. The full generation $(R, A)$ conditioned on $Q$ is modeled as:
\begin{equation}\label{eq:def rlm}
P(R, A \mid Q) = \prod_{i=1}^{|R|} P_{\theta}(r_i \mid Q, r_{<i}) \prod_{j=1}^{|A|} P_{\theta}(a_j \mid Q, R, a_{<j}),
\end{equation}
where $R=(r_1,r_2,...,r_{|R|})$ represents the sequence of reasoning tokens.

In practice, the reasoning content $R$ and the final answer $A$ are typically separated by special delimiter tokens, such as "$<think>$" and "$</think>$".

\subsection{Reasoning Behavior}

Given a reasoning sequence $R$, we characterize the underlying reasoning process as a structured sequence of behavioral units. Following the taxonomy established by \cite{Yang2025TesttimePI}, we define the set of reasoning behaviors as $\mathcal{B} = \{O, P, S, E, V, C\}$, where each element represents a distinct functional role:

\begin{itemize}
    \item \textbf{Objective ($O$)} clarifies the goal of the reasoning process and the desired final outcome. It is usually articulated at the beginning of the reasoning.
    \item \textbf{Progression ($P$)} denotes pushing forward along the current line of thought by applying known facts and inference rules, often signaled by discourse markers such as “Next”, “Then”.
    \item \textbf{Summary ($S$)} denotes consolidating and reorganizing the key information obtained so far in order to prepare for subsequent reasoning, often introduced by summarizing phrases like “In summary”.
    \item \textbf{Exploration ($E$)} denotes proposing new hypotheses or considering alternative solution strategies when the current line of reasoning stalls or appears unpromising, often cued by expressions like “Alternatively”.
    \item \textbf{Verification ($V$)} denotes examining recent reasoning steps to ensure their logical soundness and correctness, and is typically preceded by signals such as “Wait”.
    \item \textbf{Conclusion ($C$)} denotes presenting the final answer once sufficient and reliable reasoning has been established.
\end{itemize}

To analyze the reasoning process, we extract the reasoning behavior chain $\mathbf{b}$ from the raw text $R$. This process involves two steps: segmentation and classification.

We partition the reasoning text $R$ into an ordered sequence of segments $\mathcal{S} = (s_1, s_2, \dots, s_n)$ using the double-newline delimiter $d = \text{\textbackslash n\textbackslash n}$:
\begin{equation}
\mathcal{S} = \text{Split}(R, \text{delimiter}=d).
\end{equation}
For each segment $s_i \in \mathcal{S}$, we identify its behavior type $b_i \in \mathcal{B}$ via a mapping function $\phi: \mathcal{S} \rightarrow \mathcal{B}$. This mapping is determined by identifying specific \textit{marker tokens} at the onset of each segment $b_i = \phi(s_i)$, where $\phi$ is defined by the rules in \cref{app:behavior identify rules}.

By applying $\phi$ to the sequence $\mathcal{S}$, we obtain the reasoning behavior chain $\mathbb{b}$:
\begin{equation}
\mathbf{b} = (b_1, b_2, \dots, b_n), \quad b_i \in \mathcal{B}.
\end{equation}

Typically, a reasoning behavior chain $\mathbf{b}$ initiates with $O$ and terminates with $C$, traversing various instances of $\{P, S, E, V\}$ as intermediate states. In complex scenarios, $\mathbf{b}$ could exhibit a non-linear reasoning structure, such as multiple $C$ steps representing the derivation of sub-conclusions before the final synthesis.

\section{Main Method}

This paper mainly analyzes the reasoning process from the perspective of behavior patterns. \cref{fig:reasoning_dna} shows our main conceptions.
We first introduce the definition of reasoning behavior patterns. 
To analyze how changes in behavior patterns affect LLM models, we subsequently introduce a method for structurally injecting behavior patterns.
Building on the behavior pattern injection method, we propose two methods, InjectCorrect and InjectRLOpt, to optimize the model’s behavior pattern at reasoning time.

\subsection{Reasoning Behavior Pattern}
\begin{figure}
    \centering
    \includegraphics[width=\columnwidth]{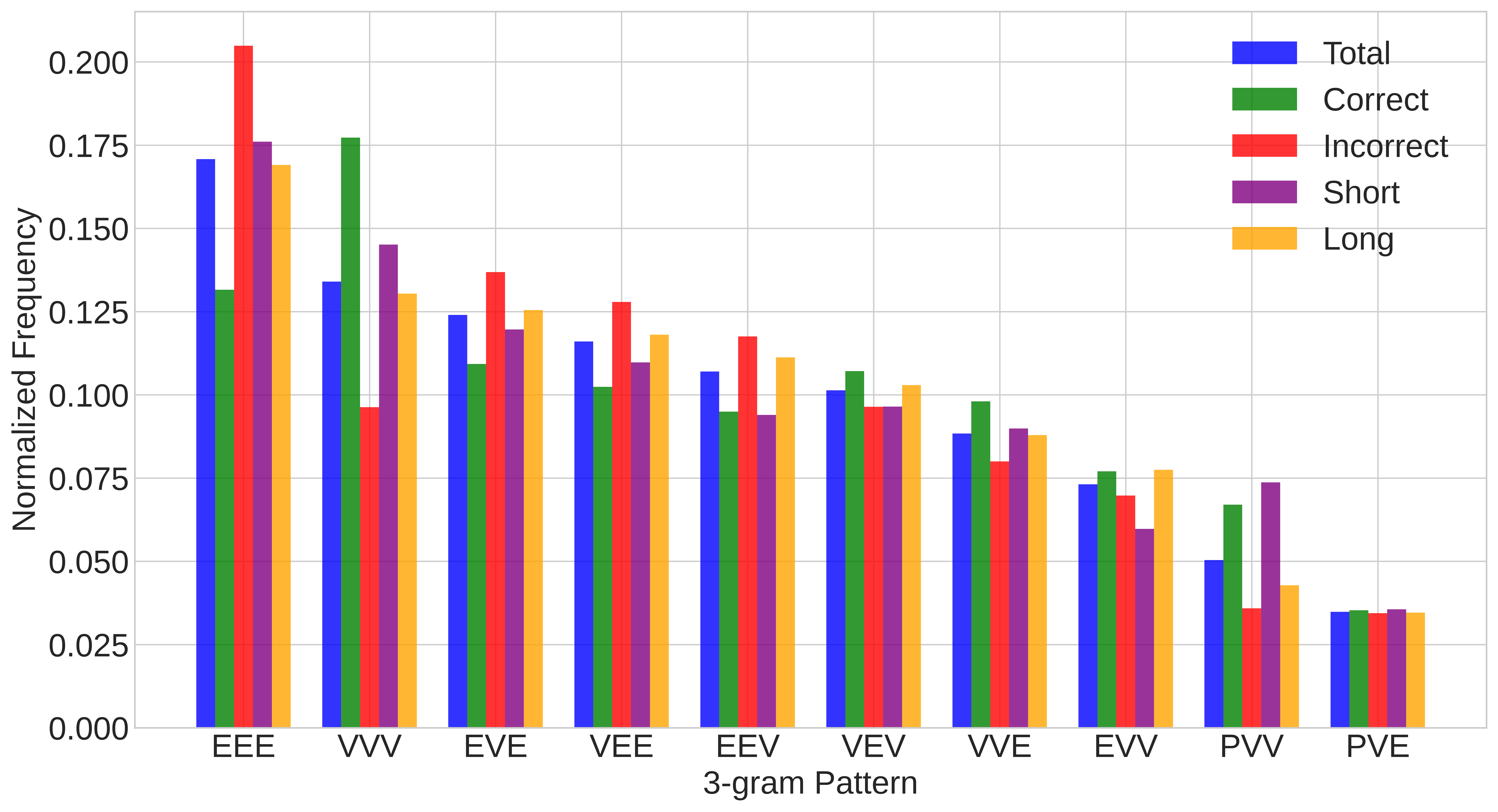}
    \caption{Normalized frequencies of the top‑10 reasoning behavior patterns of Qwen3‑8B on the GPQA dataset. For each pattern, its frequency is shown for total, correct, incorrect, short‑reasoning, and long‑reasoning samples. Within each sample subset, frequencies are normalized to sum to 1.}
    \label{fig:true_false_pattern}
\end{figure}

A reasoning behavior chain is a sequence of diverse behaviors arranged in the order of their occurrence. As shown in \cref{fig:reasoning_dna} (b) We treat each atomic-level reasoning behavior $b \in \mathcal{B} = \{O, P, V, E, S, C\}$ as a discrete symbol, analogous to the nucleotide bases (A, T, C, G) in DNA. Under this analogy, the reasoning behavior chain constitutes a sequence in which these symbols are ordered temporally or logically, similar to a nucleotide sequence. Different ways of combining and ordering these behaviors, just as different base sequences encode different proteins, give rise to distinct reasoning trajectories and conclusions.

Formally, a behavior chain $\mathbf{b}$ of length $L$ is an ordered sequence $\mathbf{b} = (b_1, b_2, \ldots, b_L)$, where $b_t \in \mathcal{B}$ denotes the behavior at step $t$. To capture local structural units, we define a length-$n$ local behavior segment (i.e., an $n$-gram) as $g \in \mathcal{G}^{(n)}$, where $\mathcal{G}^{(n)} = \mathcal{B}^n$ is the set of all possible $n$-gram patterns. 

Given a corpus of behavior chains $\mathcal{D} = \{\mathbf{b}^{(k)}\}_{k=1}^K$, where each chain $\mathbf{b}^{(k)}$ has length $L_k$, we compute the frequency of each $n$-gram pattern $g \in \mathcal{G}^{(n)}$ as:
\begin{equation}
\text{freq}_{\mathcal D}(g) = \sum_{k=1}^{K} \sum_{t=1}^{L_k - n + 1} \mathbf{1}\Big( (b^{(k)}_t, \dots, b^{(k)}_{t+n-1}) = g \Big),
\end{equation}
where $\mathbf{1}(\cdot)$ is the indicator function. These counts are then normalized to obtain the empirical distribution:
\begin{equation}
p^{(n)}(g) = \frac{\text{freq}_{\mathcal D}(g)}{\sum_{g' \in \mathcal{G}^{(n)}} \text{freq}_{\mathcal D}(g')}.
\end{equation}

Analogous to regulatory motifs in genomics, these $n$-gram behaviour patterns can be viewed as micro-strategies with specific functional or semantic roles, for example:
\begin{itemize}
    \item $OP$: The model first establishes or restates the task objective ($O$), and then immediately proceeds with step-by-step reasoning or execution ($P$). This pattern reflects goal-grounded execution, where subsequent reasoning is explicitly anchored to a clearly articulated objective.
    \item $PV$: The model carries out detailed reasoning ($P$) and then directly examines the resulting steps or conclusions for logical or computational errors ($V$). This is a derive-then-check pattern, characteristic of proof checking and numerical verification.
    \item $ESV$: The model first explores possible approaches or hypotheses ($E$), then summarizes or consolidates the intermediate findings ($S$), and finally verifies their correctness ($V$). This forms a complete micro-loop of explore → consolidate → validate, functioning as a local reflective cycle that enhances robustness and reliability.
\end{itemize}

Thus, each frequent $n$-gram is not only a statistical co-occurrence pattern, but can also be interpreted as a semantically functionally coherent local reasoning process.

\cref{fig:true_false_pattern} presents the n-gram distributions of the Qwen3-8B model’s reasoning behavior when answering questions in the GPQA dataset. We observe substantial differences in the distributions of reasoning behavior patterns between correct and incorrect answers, whereas those between longer and shorter answers differ only marginally. This suggests that reasoning behavior patterns may influence the quality of the final answers. Motivated by this observation, we aim to explore how changing reasoning behavior patterns affects the model’s final answers, and therefore propose a behavior pattern injection method to steer the model’s reasoning process, which we describe in detail in the next section.

\subsection{Structural Injection of Reasoning Behavior Patterns}

Reasoning behavior injection refers to the control of a model’s reasoning process such that, at each step, the next reasoning behavior is sampled according to a prescribed probability distribution over behavior patterns. 
This behavior injection is analogous to DNA editing, as illustrated in \cref{fig:reasoning_dna} (c), in which the final gene expression is altered by modifying specific gene segments.

Formally, at time step $t \ge 1$, given the sequence of preceding behaviors $\mathbf{b}_{<t} = (b_1, \dots, b_{t-1})$, the next behavior $b_t$ is sampled from a conditional distribution. We employ a variable-order $n$-gram scheme where a tunable hyperparameter $n \in \mathbb{N}$ specifies the maximum context length. At each step $t$, we define the effective $n$-gram order as:
\begin{equation}
    n_t = \min(n, t),
\end{equation}
which allows the model to automatically back off to shorter contexts during the early stages of reasoning ($t < n$). The context $c_t$ is defined as the sequence of the most recent $n_t - 1$ behaviors:
\begin{equation}
    c_t = (b_{t-n_t+1}, \dots, b_{t-1}),
\end{equation}
where $c_t$ is an empty sequence when $n_t = 1$ (i.e., at $t=1$). For any candidate behavior $b \in \mathcal{B}$, we consider the concatenated pattern $(c_t, b) \in \mathcal{G}^{(n_t)}$. For a given corpus $\mathcal D$, the probability of sampling $b$ is given by:
\begin{equation}
    P(b \mid c_t) = \frac{\mathrm{freq}_{\mathcal D}\bigl((c_t, b)\bigr)}{\sum_{b' \in \mathcal{B}} \mathrm{freq}_{\mathcal D}\bigl((c_t, b')\bigr)},
\end{equation}
where $\mathrm{freq}(\cdot)$ denotes the occurrence count of a pattern in the corpus of behavior chains $\mathcal{D}$.

The hyperparameter $n$ governs the trade-off between contextual expressiveness and statistical reliability.
This setting aligns closely with research on fast-and-frugal heuristics, particularly in time-series forecasting \cite{katsikopoulos2020classification}. Empirical studies have demonstrated that recency heuristics—which rely on very short contexts ($n=1$ or $2$)—often outperform complex models in autocorrelated environments \cite{KATSIKOPOULOS2022613,CASTLE2022622}. 

For reasoning behavior patterns, when $n = 2$, the control mechanism conditions only on pairwise adjacency between behaviors (bi-grams), which is generally insufficient to capture richer local substructures in reasoning. 
When $n = 3$, the context corresponds to behavior tri-grams, which are expressive enough to encode the smallest recurring structural units in reasoning patterns and to distinguish between several common local motifs. For example, 3-step patterns can already capture typical elementary reasoning units such as “progression--exploration-verification” and “exploration-exploration-summary.” 
For $n \geq 4$, the longer context allows finer-grained discrimination among behavior patterns. However, the space of possible behavior sequences grows exponentially with $n$. Consequently, many context segments $c_t$ induce patterns $(c_t, b)$ that are rarely or never observed in $\mathcal D$. Even when such patterns do appear, their counts are typically too small to support reliable estimation of $P(b \mid c_t)$, leading to high-variance behavior control.

From the perspective of behavior control, shorter context lengths (smaller $n$) yield more robust statistics for each behavior pattern at the cost of weaker contextual control, whereas longer contexts (larger $n$) encode richer structural information but suffer from fewer samples per pattern and thus higher variance in the empirical behavior distribution. Besides, the marginal benefit of extending the local structure from 3 to 4 or more steps is relatively limited, whereas the adverse impact of data sparsity becomes much more pronounced. At the scale of our corpus $\mathcal D$, 3-gram patterns already provide sufficient coverage of common elementary reasoning structures while maintaining adequate statistical support. Therefore, unless otherwise specified, we fix $n = 3$ in all experiments and perform 3-gram-based reasoning behavior injection. 

\begin{table*}[t]
  \centering
  \small
  \setlength{\tabcolsep}{5pt}
  \caption{Frequencies of Top 5 3-gram patterns across Qwen3 series models and GPQA, MATH, AIME25, MBPP datasets. Here, O: Objective, P: Progession, S: Summary, E: Exploration, V: Verification. Results are averaged over five runs.}
  \label{tab:3gram_freq}
  \begin{tabular}{l*{2}{cccc}}
    \toprule
    & \multicolumn{4}{c}{Qwen3-32B}
    & \multicolumn{4}{c}{Qwen3-14B} \\
    \cmidrule(lr){2-5}
    \cmidrule(lr){6-9}
    Rank
      & GPQA & MATH & AIME25 & MBPP
      & GPQA & MATH & AIME25 & MBPP \\
    \midrule
    \textbf{Top 1}
      & \cellcolor{Ecol!40}EEE (1.67) & \cellcolor{Vcol!40}VVV (1.04) & \cellcolor{Ccol!40}CCC (3.86) & \cellcolor{Vcol!40}VVV (1.07)
      & \cellcolor{Ecol!40}EEE (2.11) & \cellcolor{Vcol!40}VVV (1.58) & \cellcolor{Ecol!40}EEE (4.72) & \cellcolor{Vcol!40}VVV (1.42) \\
    \textbf{Top 2}
      & \cellcolor{Vcol!40}VVV (1.32) & \cellcolor{Ecol!40}EEE (0.89) & \cellcolor{Ecol!40}EEE (3.49) & \cellcolor{Ecol!40}EEE (0.53)
      & \cellcolor{Vcol!40}VVV (1.95) & \cellcolor{Ecol!40}EEE (1.21) & \cellcolor{Vcol!40}VVV (4.33) & \cellcolor{Ecol!40}EEE (0.81) \\
    \textbf{Top 3}
      & VEE (1.04) & \cellcolor{Ccol!40}CCC (0.85) & \cellcolor{Vcol!40}VVV (3.27) & PVV (0.50)
      & EVE (1.84) & \cellcolor{Pcol!40}PPP (0.85) & \cellcolor{Ccol!40}CCC (2.81) & VVE (0.52) \\
    \textbf{Top 4}
      & EVE (1.04) & \cellcolor{Pcol!40}PPP (0.80) & VCC (2.21) & VVC (0.43)
      & VEE (1.71) & VEE (0.80) & VVC (2.23) & VEV (0.50) \\
    \textbf{Top 5}
      & EEV (0.93) & VCC (0.62) & VVC (2.05) & OPV (0.42)
      & VEV (1.57) & VVE (0.71) & VCV (2.11) & VEE (0.49) \\
    \midrule
    & \multicolumn{4}{c}{Qwen3-8B}
    & \multicolumn{4}{c}{Qwen3-4B} \\
    \cmidrule(lr){2-5}
    \cmidrule(lr){6-9}
    \textbf{Top 1}
      & \cellcolor{Ecol!40}EEE (3.63) & \cellcolor{Vcol!40}VVV (1.94) & \cellcolor{Ecol!40}EEE (6.01) & \cellcolor{Vcol!40}VVV (1.98)
      & \cellcolor{Ecol!40}EEE (6.28) & \cellcolor{Ecol!40}EEE (2.61) & \cellcolor{Ecol!40}EEE (8.15) & \cellcolor{Vcol!40}VVV (1.52) \\
    \textbf{Top 2}\
      & \cellcolor{Vcol!40}VVV (2.85) & \cellcolor{Ecol!40}EEE (1.78) & \cellcolor{Vcol!40}VVV (4.60) & \cellcolor{Ecol!40}EEE (0.95)
      & EVE (2.86) & \cellcolor{Vcol!40}VVV (1.48) & \cellcolor{Vcol!40}VVV (4.21) & \cellcolor{Ecol!40}EEE (1.40) \\
    \textbf{Top 3}
      & EVE (2.64) & VEE (1.07) & \cellcolor{Ccol!40}CCC (3.53) & VEV (0.68)
      & VEE (2.81) & VEE (1.19) & \cellcolor{Ccol!40}CCC (3.21) & VEE (0.65) \\
    \textbf{Top 4}
      & VEE (2.47) & \cellcolor{Pcol!40}PPP (0.88) & VVC (2.67) & VEE (0.66)
      & EEV (2.60) & EEV (0.89) & VVC (2.37) & VEO (0.62) \\
    \textbf{Top 5}
      & EEV (2.28) & VVE (0.88) & CVV (2.60) & VVE (0.64)
      & \cellcolor{Vcol!40}VVV (1.76) & EVE (0.88) & CVV (2.34) & VEV (0.58) \\
    \bottomrule
  \end{tabular}
\end{table*}

\subsection{Reasoning Behavior Improvement}

Building on the observation that behavior-pattern injection significantly influences response quality, we propose to steer the model's reasoning process by optimizing the behavior distribution $P(b \mid c)$. We explore two methods for this optimization: InjectCorrect and InjectRLOpt. The main procecure of these methods is shown in \cref{fig:reasoning_dna} (d)

The InjectCorrect method learning a aciton distribution from the behavioral priors of successful reasoning trajectories. It could be seen as a self-imitation learning \cite{oh2018selfimitationlearning,chen2020bail}. Specifically, let $\mathcal{D}_{corr} \subset \mathcal{D}$ be the subset of reasoning chains that result in a correct final answer. We define the optimized distribution $P_{corr}(b \mid c)$ as the empirical conditional distribution derived exclusively from $\mathcal{D}_{corr}$:
\begin{equation}
    P_{corr}(b \mid c) = \frac{\mathrm{freq}_{\mathcal{D}_{corr}}((c, b))}{\sum_{b' \in \mathcal{B}} \mathrm{freq}_{\mathcal{D}_{corr}}((c, b'))},
\end{equation}
where $\mathrm{freq}_{\mathcal{D}_{corr}}$ denotes the pattern count within the \textit{correct-only} corpus. This approach steers the model toward local motifs that are statistically correlated with successful outcomes.

In the InjectRLOpt method,  we formulate the reasoning process as a Markov Decision Process (MDP) and optimize the behavior distribution using reinforcement learning \cite{sutton2018reinforcement,watkins1992q,mnih2015human}. In this framework, each state $s_t \in \mathcal{S}$ is characterized by the current $(n-1)$-gram context $c_t \in \mathcal{G}^{(n-1)}$, which captures the local history of reasoning behaviors, while an action $a_t \in \mathcal{A}$ corresponds to the selection of the subsequent behavior $b_t \in \mathcal{B}$. Transitions within the behavior space are deterministic: the next state $s_{t+1}$ is generated by appending the chosen action $a_t$ to the current context $s_t$ and shifting the $n$-gram window. To align the model's reasoning with answer correction, we employ a sparse terminal reward function $R(s_t, a_t)$ based on the correctness of the final answer $A$. The reward function is  defined as:
\begin{equation}
    R(s_t, a_t) = 
    \begin{cases} 
    1 & \text{if } t=T \text{ and } A \text{ is correct,} \\
    0 & \text{otherwise.}
    \end{cases}
\end{equation}
Based on this MDP formulation, for any reasoning trajectory $\tau = \{(s_t, a_t, r_t)\}_{t=1}^T$, we define the discounted return as $G_t = \sum_{k=t}^T \gamma^{k-t} r_k$, where $\gamma \in [0, 1]$ is the discount factor. The corresponding action-value function, $Q(s, a) = \mathbb{E}[G_t \mid s_t=s, a_t=a]$, represents the expected success rate of a behavior given its local context. In practice, we derive an empirical estimate $\hat{Q}(s, a)$ by averaging the returns of all trajectories in $\mathcal{D}$ that contain the state-action pair $(s, a)$.

However, a standard value-based Boltzmann policy \cite{reif1965statistical} may be sensitive to noise, particularly for rare behavior patterns, where the empirical estimate $\hat{Q}(s, a)$ is based on a limited number of samples. To account for this varying statistical confidence, we propose a \textit{Reliability-Aware Softmax Policy}. Instead of relying solely on the estimated Q-values, this policy modulates the sampling probability using a visitation-based confidence weight:
\begin{equation}
    \pi(a \mid s) = \frac{\exp(\hat{Q}(s, a) / \tau) \cdot f(N(s, a))}{\sum_{a' \in \mathcal{B}} \exp(\hat{Q}(s, a') / \tau) \cdot f(N(s, a'))},
\end{equation}
where $\tau > 0$ is the temperature parameter and $N(s, a)$ denotes the visitation count of the pair $(s, a)$ in $\mathcal{D}$. The confidence weight $f(N)$ is defined as:
\begin{equation}
    f(N) = \frac{\sqrt{N}}{\sqrt{N} + c},
\end{equation}
where $c > 0$ is a hyperparameter that controls the penalty for low-frequency patterns, we set $c = 10$ in our experiments. This mechanism serves as a frequentist regularizer that down-weights $Q$-values with high epistemic uncertainty, thereby steering the model toward reasoning motifs with robust statistical evidence. As the data density increases ($N(s, a) \to \infty$), the factor $f(N)$ approaches $1$, and the policy naturally recovers the standard Boltzmann distribution \cite{reif1965statistical}.

\section{Experiments}

\begin{table*}[t]
    \centering
    \caption{Performance (mean $\pm$ std) of Qwen3 series models on four datasets (GPQA, MATH, AIME25, MBPP) under different methods for adjusting the distribution of reasoning behavior patterns. Results are averaged over five runs.}
    \label{tab:performance_improve}

    \newcommand{\res}[2]{#1\hspace{1pt}\textcolor{gray}{\tiny{$\pm$#2}}}

    \setlength{\tabcolsep}{1.5pt} 

    \resizebox{\textwidth}{!}{
        \begin{tabular}{l *{8}{c}}
            \toprule
            & \multicolumn{4}{c}{Qwen3-32B} 
            & \multicolumn{4}{c}{Qwen3-14B} \\
            \cmidrule(lr){2-5}
            \cmidrule(lr){6-9}
            Method 
            & GPQA & MATH & AIME25 & MBPP
            & GPQA & MATH & AIME25 & MBPP \\
            \midrule
            
            Baseline
            & \res{0.6061}{0.022} & \res{0.9384}{0.003} & \res{0.6933}{0.043} & \res{0.5876}{0.004}
            & \res{0.6040}{0.015} & \res{\textbf{0.9364}}{0.004} & \res{0.5933}{0.086} & \res{\textbf{0.5744}}{0.004} \\
            
            InjectCorrect 
            & \res{0.6232}{0.017} & \res{0.9380}{0.003} & \res{0.7000}{0.100} & \res{\textbf{0.5916}}{0.009}
            & \res{0.6051}{0.013} & \res{0.9296}{0.006} & \res{0.6467}{0.069} & \res{0.5648}{0.005} \\
            
            InjectRLOpt 
            & \res{\textbf{0.6444}}{0.027} & \res{\textbf{0.9416}}{0.003} & \res{\textbf{0.7111}}{0.040} & \res{0.5908}{0.004}
            & \res{\textbf{0.6333}}{0.017} & \res{0.9332}{0.004} & \res{\textbf{0.6733}}{0.049} & \res{0.5704}{0.003} \\
            
            \midrule
            & \multicolumn{4}{c}{Qwen3-8B}
            & \multicolumn{4}{c}{Qwen3-4B} \\
            \cmidrule(lr){2-5}
            \cmidrule(lr){6-9}
            
            Baseline
            & \res{0.5869}{0.020} & \res{0.9252}{0.005} & \res{0.5867}{0.044} & \res{0.5604}{0.008}
            & \res{0.5222}{.008} & \res{0.9288}{0.004} & \res{0.5533}{0.038} & \res{0.5520}{0.004} \\
            
            InjectCorrect 
            & \res{0.5707}{0.024} & \res{0.9233}{0.007} & \res{0.6133}{0.050} & \res{0.5612}{0.004}
            & \res{\textbf{0.5414}}{0.014} & \res{0.9240}{0.005} & \res{0.5733}{0.043}  & \res{0.5523}{0.009} \\
            
            InjectRLOpt 
            & \res{\textbf{0.5949}}{0.018} & \res{\textbf{0.9328}}{0.004} & \res{\textbf{0.6600}}{0.054} & \res{\textbf{0.5652}}{0.008}
            & \res{0.5323}{0.029} & \res{\textbf{0.9288}}{0.003} & \res{\textbf{0.6400}}{0.037} & \res{\textbf{0.5580}}{0.005} \\
            
            \bottomrule
        \end{tabular}
    }
\end{table*}

Our experiments are designed to investigate the structural and functional properties of reasoning behaviors by addressing three key objectives:
\begin{itemize}
    \item \textbf{RQ1 (Domain Specificity):} Does the model exhibit consistent and identifiable reasoning behavior patterns (motifs) that are characteristically associated with successful problem-solving within specific domains?
    \item \textbf{RQ2 (Pattern Transferability):} Are these behavior patterns domain-dependent? Specifically, does the injection of behavior patterns derived from unrelated problem categories lead to a subsequent degradation in answer quality?
    \item \textbf{RQ3 (Behavioral Engineering):} To what extent can we improve the model’s success rate purely by injecting optimized behavior distributions without modifying the underlying model parameters?
\end{itemize}

\subsection{Setup}
\textbf{Models.} To investigate how reasoning behaviors evolve across different model scales, we employ the Qwen3 series \cite{qwen3technicalreport}, including the 32B, 14B, 8B, and 4B variants. We specifically select this family because these models generate explicit, human-readable reasoning traces (Chain-of-Thought). This transparency is essential for our methodology, as it allows us to both extract behavior chains from raw outputs and perform granular interventions by injecting behavior marker tokens to steer the reasoning process.

\textbf{Datasets.} Our evaluation spans four demanding reasoning-intensive benchmarks to ensure the generalizability of our findings: GPQA \cite{rein2024gpqa}, AIME25\cite{aime25}, MATH \cite{lightman2023lets}, and MBPP \cite{austin2021program}. Collectively, these datasets provide a comprehensive testbed covering scientific inquiry, mathematical deduction, and code generation.

\subsection{Main Results}

\textbf{Large language models' reasoning behavior patterns adapt to the characteristics of problems it solves.} \cref{tab:3gram_freq} presents the Top 5 3-gram behavior patterns and their corresponding frequencies for the Qwen3 models across multiple datasets. The frequency is computed as the total number of occurrences of a given pattern across all reasoning instances divided by the total number of reasoning instances, i.e., the average number of times that pattern appears in a single reasoning instance.
\cref{tab:3gram_freq} shows that, across different tasks, EEE and VVV are among the model’s most frequently used behavior patterns. This observation aligns with prior research highlighting the efficacy of extensive sampling or exploration \cite{wang2023selfconsistencyimproveschainthought,tot-2023} and the critical role of iterative self-verification and refinement \cite{lightman2023lets,madaan2023selfrefine} in complex reasoning tasks. At the same time, the distributions of behavior patterns still differ significantly across tasks. Specifically, on the GPQA dataset, the model more frequently adopts exploration-related patterns, whereas on the mathematical datasets MATH and AIME25, it more often employs stepwise patterns (e.g., PPP, CCC, VVC).

Intuitively, this difference aligns with the human reasoning demands of the tasks. GPQA questions typically draw on diverse knowledge sources, so the model tends to “branch out, probe, and then converge,” \cite{guilford1967nature,cropley2006praise} leading to repeated trial-and-error and exploratory behavior patterns, which is a tree-like reasoning process. In contrast, the structured nature of MATH and AIME25 tasks corresponds to the psychological classification of well-defined problems \cite{1964Heuristic, Simon1977}, with relatively linear and decomposable solution paths \cite{newell1972human} . As a result, the model is more inclined to follow a chain-like trajectory, incrementally accumulating and verifying partial conclusions, which manifests as stable, stepwise patterns such as PPP, CCC, and VVC.

\begin{table}[ht]
    \centering
    \caption{The accuracy of Qwen3-8B on the GPQA dataset when specific patterns are masked during reasoning. Each column is a prefix, and each row corresponds to a masked action. Results are averaged over five runs. Qwen3-8B base accuracy: 0.5869}
    \label{tab:8b_gpqa_rm_pattern}
    \setlength{\tabcolsep}{8pt}
    \begin{tabular}{lcccc}
        \toprule
        & \textbf{EE} & \textbf{EV} & \textbf{PE}\\
        \midrule
        \textbf{E}xploration & 0.5848 & 0.5828 & \textbf{\textit{0.5960}}\\
        \textbf{V}erification & 0.5869 & \textbf{\textit{0.5919}} & 0.5859 \\
        \textbf{P}rogression & \textbf{\textit{0.5970}} & 0.5758 & \textbf{0.5631}\\
        \textbf{C}onclusion & 0.5889 & \textbf{0.5606} & 0.5896\\
        \bottomrule
    \end{tabular}
\end{table}

\textbf{Specific reasoning behavior patterns can emerge that substantially influence their reasoning outcomes.}
We conduct experiments with Qwen3-8B on the GPQA dataset, masking certain 3-gram behavior patterns to examine how they affect the quality of the model’s final answers. Specifically, for a given 3-gram behavior pattern, once its 2-gram prefix is detected, we remove the last action of that 3-gram from the candidate action set, thereby masking the corresponding 3-gram behavior pattern.
\cref{tab:8b_gpqa_rm_pattern} reports experimental results when certain patterns are masked during reasoning. From this, we observe that masking certain behavior patterns (e.g., EEP, PEE, EVV) increases the model’s final accuracy. In contrast, masking another set of behavior patterns (e.g., EVC, PEP) causes a substantial drop in accuracy, while masking the remaining patterns does not result in any significant change in the model’s final accuracy. 

A possible explanation for this phenomenon is that some behavior patterns can be substituted with alternative reasoning paths, so masking them has little effect on reasoning quality. By contrast, other behavior patterns are harder to replace, and masking them could substantially influence the model’s reasoning quality. Therefore, we emphasize the critical role of reasoning behavior patterns in the reasoning process and argue that adjusting these patterns in large language models can steer their reasoning and ultimately influence the quality of final answers.

\begin{table}[ht]
    \centering
    \caption{
        Cross-injection reasoning pattern transfer accuracy of Qwen3-8B between datasets.
        Patterns are extracted from the row dataset and injected into the column dataset.
        Results are averaged over five runs.
    }
    \label{tab:dataset_transfer}
    \setlength{\tabcolsep}{8pt}
    \begin{tabular}{lcccc}
        \toprule
        & GPQA & MATH & AIME25 & MBPP \\
        \midrule
        GPQA   & /      & 0.9220 & 0.5778 & 0.5620 \\
        MATH   & \textbf{0.5879} &  /     & \textbf{0.6333} & \textbf{0.5630} \\
        AIME25 & 0.5798 & \textbf{0.9340} &  /     & 0.5560 \\
        MBPP   & 0.5596 & 0.9320 & 0.5667 &  /     \\
        \bottomrule
    \end{tabular}
\end{table}

\textbf{Large language models exhibit distinct reasoning patterns across different types of problems, and these patterns are adapted to the corresponding problem types.}
Table \ref{tab:dataset_transfer} provides empirical support for this hypothesis by illustrating the transfer accuracy of reasoning patterns across various tasks. The results demonstrate that the effectiveness of pattern injection is highly sensitive to the semantic alignment between the extraction source and the injection target. 
We consistently observe that reasoning patterns transferred between homogeneous tasks yield superior performance compared to those transferred across heterogeneous domains.
This task-specific adaptation is most prominent in the mathematical domain. As indicated by the results for AIME25, a challenging mathematics dataset, injecting reasoning patterns extracted from the MATH dataset results in an accuracy of 0.6333. This performance significantly surpasses that of patterns derived from disparate domains, such as general science (GPQA, 0.5778) or code generation (MBPP, 0.5667). 

Conversely, significant domain shifts lead to suboptimal transfer outcomes. For the MBPP dataset, which focuses on code synthesis, injecting reasoning behavior patterns from MATH or GPQA results in stagnant performance (ranging from 0.55 to 0.56), indicating that the algorithmic logic required for coding is distinct from the chain-of-thought patterns used in arithmetic or general knowledge reasoning. These findings collectively imply that reasoning in LLMs is not a monolithic process that can be universally applied; rather, it is a structured, multidimensional process in which the reasoning structure must match the intrinsic requirements of the target problem.

\textbf{Adjusting large language models’ reasoning pattern distribution can improve their final performance.} \cref{tab:performance_improve} reports the accuracy of the Qwen3 model on different datasets under two reasoning behavior pattern distribution injection methods, InjectCorrect and InjectRLOpt. InjectCorrect is a method that demonstrates injecting the behavior pattern distribution extracted from the models' reasoning process with correct answers. InjectRLOpt is a method that injects a behavior pattern distribution optimized by a reinforcement learning algorithm. Specifically, it learns a value function from the models’ reasoning behavior chains and correct/incorrect labels of their answers.
From a practical perspective, we compute a Boltzmann policy from the value function as the behavior pattern distribution injectant.

From the experimental results in \cref{tab:performance_improve}, InjectRLOpt generally achieves the optimal overall performance: for most models and datasets, it matches or exceeds the baseline and also outperforms InjectCorrect in many cases. InjectCorrect yields modest gains over the baseline on a subset of tasks. Across different model sizes, the relative performance ranking of the three methods is largely consistent, suggesting that the benefits of this approach can transfer across scales.

\subsection{Hyperparameter Choice}

As illustrated in Table \ref{tab:gpqa_results}, we observe that for larger models (32B, 14B, and 8B), the reasoning performance peaks at $\gamma = 0.98$, suggesting that a moderate discount factor effectively balances the temporal credit between intermediate reasoning steps and the final terminal reward. In contrast, the smallest variant (4B) achieves its best result at $\gamma = 1.0$. Given its robust performance across most scales, we recommend $\gamma = 0.98$ as the default hyperparameter for optimizing reasoning behavior by the InjectRLOpt method.

\begin{table}[tbp]
\centering
\caption{Impact of the discount factor $\gamma$ on GPQA performance across various model scales. Results are averaged over five runs.}
\label{tab:gpqa_results}
\begin{tabular}{lcccc}
\toprule
 & \multicolumn{4}{c}{Qwen3} \\
\cmidrule{2-5}
Method & 32B & 14B & 8B & 4B \\
\midrule
base          & 0.6061 & 0.6040 & 0.5869 & 0.5222 \\
$\gamma = 1$    & 0.6343 & 0.6141 & 0.5859 & \textbf{0.5394} \\
$\gamma = 0.99$ & 0.6293 & 0.6230 & 0.5899 & 0.5303 \\
$\gamma = 0.98$ & \textbf{0.6444} & \textbf{0.6333} & \textbf{0.5949}  & 0.5323 \\
$\gamma = 0.96$ & 0.6414 & 0.6020 & 0.5899 & 0.5242 \\
\bottomrule
\end{tabular}
\end{table}

\section{Results and Conclusions }

This paper analyzes the reasoning behavior patterns of large language models. We find that large language models exhibit adaptive behavior patterns across different types of tasks, and that changes in these patterns can significantly affect their reasoning process and answer quality. Building on these findings, we propose two methods for improving reasoning behavior patterns: InjectCorrect and InjectRLOpt. By modifying the model’s behavior patterns during reasoning without changing its parameters, these methods substantially enhance the model’s reasoning quality and answer accuracy.

Due to computational constraints, our experiments and analyses only consider 3‑gram behavior patterns. Future work can investigate more diverse behavior patterns. Moreover, this paper only examines how adjusting behavior patterns, without additional training, affects the model’s answer performance. Future work may employ post-training techniques, such as supervised fine-tuning and reinforcement learning, to explicitly optimize the model’s reasoning behavior and further improve its reasoning capabilities.

\bibliographystyle{icml2026}
\bibliography{ref}

\newpage
\appendix
\onecolumn
\section{Behavior Identify Rules}
\label{app:behavior identify rules}

\begin{table}[h]
\centering
\caption{Control phrases to identify reasoning behaviors}
\label{tab:control-words}
\begin{tabular}{l|p{0.75\textwidth}}
\hline
\textbf{Category} & \textbf{Phrases} \\
\hline
Objective &
Okay,; We need to \\
\hline
Progression &
Now; First,; Next,; Hmm,; Looking at; Let me compute; Let me calculate; So, first, ; Let's compute; Let's consider; Let's think \\
\hline
Summary &
In summary \\
\hline
Verification &
Let me verify; But let me check; But wait; Wait; But maybe I should check; But just to make sure; Let me re-examine; Let me check; But let me just make sure; But let's check \\
\hline
Exploration &
Alternatively; But perhaps; Hmm. I remember; Another idea:; We might; Maybe; What if; Another approach; Another possibility \\
\hline
Conclusion &
Therefore,; **Final Answer**; Final Answer; So, in conclusion,; In conslusion; So the answer is; Hence; So, answer is; So answer is; So Answer: ; So the correct answer is; Thus, the correct answer is; I'll select option ; So yes, answer ; So the answer should be; So the answer must be; Thus we have; Thus answer \\
\hline
\end{tabular}
\end{table}

This section describes the rules used to identify behavior types in the model’s reasoning text and to extract control nodes accordingly. The overall pipeline, shown in \cref{alg:get-control-nodes}, consists of three stages: segmentation, behavior classification, and node aggregation.

In the segmentation stage, the raw reasoning text \(R\) is split using a double newline (`\textbackslash n\textbackslash n`) as the delimiter. This yields an ordered sequence of segments
\[
\mathcal{S} = (s_1, s_2, \dots, s_n) = \text{Split}(R, d),
\]
where \(d\) is the double-newline delimiter. Each segment \(s_i\) corresponds to a paragraph-like unit or a logically coherent chunk in the model’s reasoning.

Next, in the behavior classification stage, we assign a behavior label to each segment based on explicit control phrases. Concretely, for each segment \(s_i\), we apply a mapping
\[
b_i = \phi(s_i) \in \mathcal{B},
\]
where \(\mathcal{B}\) is the set of behavior categories, and \(\phi(\cdot)\) is a marker-based mapping function. The function \(\phi\) searches for predefined control phrases (listed in \cref{tab:control-words}) and maps the segment to the corresponding behavior category when a match is found. If no control phrase is detected, the behavior label is left undefined for that segment.

By applying \(\phi\) to every segment, we obtain an initial behavior sequence
\[
\mathbf{b} = (b_1, b_2, \dots, b_n), \quad b_i \in \mathcal{B} \cup \{\text{undefined}\}.
\]

Finally, in the node aggregation stage, we combine behavior labels and text segments into an ordered list of control nodes \(\mathcal{N}\). The aggregation obeys the following rules:

1. We maintain the current behavior \(b^{(\text{cur})}\) and the current aggregated segment \(\tilde{s}^{(\text{cur})}\). At initialization, the current behavior is set to \textbf{Objective}, i.e., \(b^{(\text{cur})} \leftarrow O\), and the current text is empty, \(\tilde{s}^{(\text{cur})} \leftarrow \emptyset\).

2. We then iterate through segments \(s_i\) and their labels \(b_i\) in order:
   - If \(b_i\) is undefined (no control phrase detected), we treat this segment as a continuation of the current behavior and append it to \(\tilde{s}^{(\text{cur})}\) via `ConcatSentences`:
     \[
     \tilde{s}^{(\text{cur})} \leftarrow \text{ConcatSentences}\big(\tilde{s}^{(\text{cur})}, s_i\big).
     \]
   - If \(b_i\) is defined (a new behavior is explicitly signaled), then:
     1. If \(\tilde{s}^{(\text{cur})}\) is non-empty, we append the current node \(\big(b^{(\text{cur})}, \tilde{s}^{(\text{cur})}\big)\) to \(\mathcal{N}\).
     2. We reset the current behavior and text to
        \[
        b^{(\text{cur})} \leftarrow b_i, \quad \tilde{s}^{(\text{cur})} \leftarrow s_i.
        \]

3. After processing all segments, we append the final node \(\big(b^{(\text{cur})}, \tilde{s}^{(\text{cur})}\big)\) to \(\mathcal{N}\).

This procedure produces an ordered list of control nodes
\[
\mathcal{N} = \big((b^{(1)}, \tilde{s}^{(1)}), \dots, (b^{(K)}, \tilde{s}^{(K)})\big),
\]
where each node is associated with a behavior type (e.g., Objective, Progression, Verification, Exploration, Summary, or Conclusion) and the corresponding text span. This structured representation is then used for downstream analysis and editing of the model’s reasoning process.

\begin{algorithm}[ht]
\caption{Extraction of Control Nodes from Reasoning Text}
\label{alg:get-control-nodes}
\begin{algorithmic}[1]

\REQUIRE Raw reasoning text $R$
\ENSURE Ordered list of control nodes $\mathcal{N} = \big((b^{(1)}, \tilde{s}^{(1)}), \dots, (b^{(K)}, \tilde{s}^{(K)})\big)$

\STATE // \textbf{Segmentation.}
\STATE Let the delimiter be the double newline $d = \text{\textbackslash n\textbackslash n}$.
\STATE Partition $R$ into an ordered sequence of segments
$
\mathcal{S} = (s_1, s_2, \dots, s_n) = \text{Split}(R, \text{delimiter}=d).$

\STATE // \textbf{Behavior classification.}
\FOR{$i = 1$ to $n$}
    \STATE Assign a preliminary behavior label $b_i = \phi(s_i) \in \mathcal{B},$
    where $\phi$ is the marker-based mapping defined in Appendix~\ref{app:behavior identify rules}
    (using the control phrase dictionary in \cref{tab:control-words}).
\ENDFOR
\STATE This yields an initial behavior sequence $\mathbf{b} = (b_1, b_2, \dots, b_n), \quad b_i \in \mathcal{B}.$

\STATE // \textbf{Node aggregation.}
\STATE Initialize an empty list of nodes $\mathcal{N}$.
\STATE Initialize the current behavior $b^{(\text{cur})} \leftarrow O$ (Objective)
and the current aggregated segment $\tilde{s}^{(\text{cur})} \leftarrow \emptyset$.
\FOR{$i = 1$ to $n$}
    \IF{$b_i$ is undefined (i.e., no behavior marker detected)}
        \STATE Concatenate $s_i$ to $\tilde{s}^{(\text{cur})}$: $ \tilde{s}^{(\text{cur})} \leftarrow \text{ConcatSentences}\big(\tilde{s}^{(\text{cur})}, s_i\big).$
    \ELSE
        \IF{$\tilde{s}^{(\text{cur})} \neq \emptyset$}
            \STATE Append node $\big(b^{(\text{cur})}, \tilde{s}^{(\text{cur})}\big)$ to $\mathcal{N}$.
        \ENDIF
        \STATE Reset the current node: $
        b^{(\text{cur})} \leftarrow b_i, \quad \tilde{s}^{(\text{cur})} \leftarrow s_i.
        $
    \ENDIF
\ENDFOR
\STATE Append the final node $\big(b^{(\text{cur})}, \tilde{s}^{(\text{cur})}\big)$ to $\mathcal{N}$.

\STATE \textbf{return} $\mathcal{N}$.

\end{algorithmic}
\end{algorithm}

\newpage
\section{N-grams Statistics}

In this section, we present the top-20 most frequent 2-gram, 3-gram, and 4-gram reasoning behavior patterns used by Qwen3 models of varying sizes (32B, 14B, 8B, 4B) across multiple datasets (GPQA, MATH, AIME25, MBPP).

\begin{table*}[ht]
  \centering
  \small
  \setlength{\tabcolsep}{5pt}
  \caption{Frequencies of Top 20 2-gram patterns across models and datasets.}
  \label{tab:2gram}
  \begin{tabular}{l*{2}{cccc}}
    \toprule
    & \multicolumn{4}{c}{Qwen3-32B}
    & \multicolumn{4}{c}{Qwen3-14B} \\
    \cmidrule(lr){2-5}
    \cmidrule(lr){6-9}
    Rank
      & GPQA & MATH & AIME25 & MBPP
      & GPQA & MATH & AIME25 & MBPP \\
    \midrule
    \textbf{Top 1}
      & \cellcolor{Ecol!40}EE (3.22) & \cellcolor{Ccol!40}CC (2.47) & \cellcolor{Ccol!40}CC (8.59) & \cellcolor{Vcol!40}VV (2.21)
      & \cellcolor{Ecol!40}EE (4.37) & \cellcolor{Vcol!40}VV (3.18) & \cellcolor{Vcol!40}VV (8.56) & \cellcolor{Vcol!40}VV (2.60) \\
    \textbf{Top 2}
      & \cellcolor{Vcol!40}VV (2.96) & \cellcolor{Vcol!40}VV (2.24) & \cellcolor{Vcol!40}VV (6.96) & PV (1.07)
      & \cellcolor{Vcol!40}VV (4.14) & \cellcolor{Ecol!40}EE (2.54) & \cellcolor{Ecol!40}EE (8.47) & \cellcolor{Ecol!40}EE (1.39) \\
    \textbf{Top 3}
      & VE (2.42) & \cellcolor{Pcol!40}PP (1.98) & \cellcolor{Ecol!40}EE (6.35) & VE (1.03)
      & VE (3.87) & \cellcolor{Pcol!40}PP (2.07) & \cellcolor{Ccol!40}CC (6.90) & VE (1.18) \\
    \textbf{Top 4}
      & EV (2.13) & \cellcolor{Ecol!40}EE (1.88) & VC (5.26) & \cellcolor{Ecol!40}EE (1.01)
      & EV (3.53) & VE (2.03) & VC (5.67) & EV (1.00) \\
    \textbf{Top 5}
      & \cellcolor{Pcol!40}PP (1.91) & VC (1.58) & CV (5.05) & VC (0.86)
      & PV (1.89) & \cellcolor{Ccol!40}CC (1.91) & CV (5.23) & PV (0.83) \\
    \textbf{Top 6}
      & PV (1.82) & CV (1.56) & \cellcolor{Pcol!40}PP (4.45) & EV (0.80)
      & \cellcolor{Pcol!40}PP (1.61) & CV (1.58) & \cellcolor{Pcol!40}PP (4.63) & VC (0.69) \\
    \textbf{Top 7}
      & VP (1.25) & VE (1.43) & PC (3.87) & \cellcolor{Pcol!40}PP (0.68)
      & VP (1.24) & VC (1.57) & VE (3.97) & VP (0.47) \\
    \textbf{Top 8}
      & PE (1.07) & PC (1.21) & CP (3.43) & VP (0.67)
      & PE (1.11) & EV (1.51) & EV (3.77) & OP (0.46) \\
    \textbf{Top 9}
      & EP (0.97) & EC (1.21) & VE (2.77) & OP (0.55)
      & EP (0.96) & PV (1.35) & PC (3.33) & OV (0.35) \\
    \textbf{Top 10}
      & OP (0.89) & PV (1.16) & PV (2.64) & OV (0.41)
      & OP (0.89) & EC (1.25) & CP (3.09) & \cellcolor{Pcol!40}PP (0.32) \\
    \textbf{Top 11}
      & VC (0.83) & EV (1.02) & EV (2.56) & EC (0.38)
      & VC (0.77) & PC (1.08) & PV (2.97) & EC (0.28) \\
    \textbf{Top 12}
      & EC (0.52) & OP (0.91) & EC (2.38) & CV (0.29)
      & EC (0.63) & VP (0.93) & EC (2.62) & CV (0.16) \\
    \textbf{Top 13}
      & PC (0.46) & VP (0.80) & VP (2.25) & PC (0.26)
      & CV (0.38) & OP (0.90) & VP (2.34) & PE (0.15) \\
    \textbf{Top 14}
      & CV (0.46) & CP (0.74) & CE (2.03) & EP (0.25)
      & PC (0.29) & PE (0.71) & CE (2.33) & EP (0.13) \\
    \textbf{Top 15}
      & CP (0.24) & CE (0.72) & PE (2.01) & PE (0.25)
      & CP (0.20) & CE (0.67) & PE (2.13) & PC (0.12) \\
    \textbf{Top 16}
      & \cellcolor{Ccol!40}CC (0.16) & PE (0.67) & EP (1.89) & \cellcolor{Ccol!40}CC (0.13)
      & \cellcolor{Ccol!40}CC (0.14) & CP (0.66) & EP (2.01) & CE (0.06) \\
    \textbf{Top 17}
      & CE (0.12) & EP (0.61) & OP (1.05) & CP (0.11)
      & CE (0.13) & EP (0.66) & OP (1.03) & CP (0.05) \\
    \textbf{Top 18}
      & OV (0.11) & OV (0.08) & PO (0.11) & CE (0.11)
      & OV (0.09) & OV (0.08) & EO (0.11) & \cellcolor{Ccol!40}CC (0.04) \\
    \textbf{Top 19}
      & OE (0.03) & OC (0.04) & OC (0.09) & OE (0.05)
      & OE (0.03) & OE (0.03) & OE (0.08) & OE (0.03) \\
    \textbf{Top 20}
      & PO (0.02) & OE (0.03) & CO (0.08) & OC (0.01)
      & OC (0.01) & OC (0.02) & OV (0.06) & PO (0.01) \\
    \midrule
    & \multicolumn{4}{c}{Qwen3-8B}
    & \multicolumn{4}{c}{Qwen3-4B} \\
    \cmidrule(lr){2-5}
    \cmidrule(lr){6-9}
    Rank
      & GPQA & MATH & AIME25 & MBPP
      & GPQA & MATH & AIME25 & MBPP \\
    \midrule
    \textbf{Top 1}
      & \cellcolor{Ecol!40}EE (6.89) & \cellcolor{Vcol!40}VV (3.90) & \cellcolor{Ecol!40}EE (10.45) & \cellcolor{Vcol!40}VV (3.37)
      & \cellcolor{Ecol!40}EE (10.09) & \cellcolor{Ecol!40}EE (4.62) & \cellcolor{Ecol!40}EE (13.05) & \cellcolor{Vcol!40}VV (2.78) \\
    \textbf{Top 2}
      & \cellcolor{Vcol!40}VV (5.81) & \cellcolor{Ecol!40}EE (3.52) & \cellcolor{Vcol!40}VV (9.89) & \cellcolor{Ecol!40}EE (1.70)
      & VE (5.17) & \cellcolor{Vcol!40}VV (3.18) & \cellcolor{Vcol!40}VV (8.85) & \cellcolor{Ecol!40}EE (2.19) \\
    \textbf{Top 3}
      & VE (5.42) & VE (2.61) & \cellcolor{Ccol!40}CC (8.43) & VE (1.57)
      & EV (4.77) & VE (2.61) & \cellcolor{Ccol!40}CC (8.01) & VE (1.47) \\
    \textbf{Top 4}
      & EV (4.95) & \cellcolor{Ccol!40}CC (2.48) & VC (6.41) & EV (1.33)
      & \cellcolor{Vcol!40}VV (3.95) & \cellcolor{Ccol!40}CC (2.15) & VC (5.71) & EV (1.27) \\
    \textbf{Top 5}
      & PV (2.41) & \cellcolor{Pcol!40}PP (2.15) & CV (6.25) & PV (0.98)
      & PV (1.66) & EV (1.94) & CV (5.65) & PV (0.90) \\
    \textbf{Top 6}
      & \cellcolor{Pcol!40}PP (1.93) & CV (2.10) & VE (4.99) & VC (0.85)
      & PE (1.58) & \cellcolor{Pcol!40}PP (1.84) & VE (4.87) & VC (0.84) \\
    \textbf{Top 7}
      & VP (1.70) & EV (1.92) & \cellcolor{Pcol!40}PP (4.87) & OP (0.58)
      & \cellcolor{Pcol!40}PP (1.37) & CV (1.76) & \cellcolor{Pcol!40}PP (4.37) & OP (0.61) \\
    \textbf{Top 8}
      & PE (1.46) & VC (1.90) & EV (4.55) & VP (0.48)
      & EP (1.36) & EC (1.64) & EV (4.35) & VP (0.40) \\
    \textbf{Top 9}
      & EP (1.33) & EC (1.59) & PC (3.59) & OV (0.39)
      & VP (1.05) & VC (1.57) & PC (3.56) & EC (0.38) \\
    \textbf{Top 10}
      & VC (0.94) & PV (1.49) & CP (3.27) & EC (0.37)
      & OP (0.93) & PV (1.25) & EC (3.31) & OV (0.36) \\
    \textbf{Top 11}
      & OP (0.93) & PC (1.22) & EC (2.94) & \cellcolor{Pcol!40}PP (0.35)
      & EC (0.86) & PC (1.04) & CP (3.19) & \cellcolor{Pcol!40}PP (0.23) \\
    \textbf{Top 12}
      & EC (0.81) & VP (1.07) & PV (2.69) & CV (0.21)
      & VC (0.72) & OP (0.91) & CE (2.71) & PE (0.22) \\
    \textbf{Top 13}
      & CV (0.65) & OP (0.90) & CE (2.40) & PE (0.18)
      & CV (0.46) & CE (0.89) & PV (2.55) & CV (0.18) \\
    \textbf{Top 14}
      & PC (0.33) & CE (0.89) & VP (2.13) & EP (0.16)
      & PC (0.24) & PE (0.85) & PE (2.36) & EP (0.17) \\
    \textbf{Top 15}
      & \cellcolor{Ccol!40}CC (0.25) & EP (0.81) & PE (2.03) & PC (0.10)
      & CE (0.22) & VP (0.83) & EP (2.30) & CE (0.09) \\
    \textbf{Top 16}
      & CP (0.25) & PE (0.81) & EP (1.98) & CE (0.08)
      & CP (0.15) & EP (0.80) & VP (1.98) & PC (0.08) \\
    \textbf{Top 17}
      & CE (0.19) & CP (0.76) & OP (1.04) & \cellcolor{Ccol!40}CC (0.05)
      & \cellcolor{Ccol!40}CC (0.15) & CP (0.61) & OP (1.06) & \cellcolor{Ccol!40}CC (0.06) \\
    \textbf{Top 18}
      & OV (0.06) & OV (0.09) & PO (0.10) & OE (0.03)
      & OV (0.06) & OV (0.08) & CO (0.09) & OE (0.04) \\
    \textbf{Top 19}
      & OE (0.03) & OE (0.03) & OV (0.10) & OE (0.03)
      & OE (0.03) & OE (0.04) & OC (0.08) & CP (0.03) \\
    \textbf{Top 20}
      & PO (0.01) & OC (0.03) & OE (0.09) & OC (0.01)
      & PO (0.01) & OC (0.02) & PO (0.06) & OC (0.01) \\
    \bottomrule
  \end{tabular}
\end{table*}

\cref{tab:2gram,tab:3gram,tab:4gram} report the top-20 most frequent 2‑gram, 3‑gram, and 4‑gram patterns of reasoning behaviors across Qwen3 models of different sizes (32B, 14B, 8B, 4B) and datasets (GPQA, MATH, AIME25, MBPP). Each pattern is defined over the categorical states Objective (O), Progression (P), Summary (S), Exploration (E), Verification (V), and Conclusion (C), and the tables list their relative frequencies for each model–dataset pair. Color-highlighted marks patterns dominated by Exploration (E), Verification (V), Conclusion (C), or Progression (P).

Across all n‑gram orders and model sizes, we observe that repetitive Exploration and Verification patterns (e.g., EE, VV, EEE, VVV, EEEE, VVVV) dominate the top positions. On GPQA, the most frequent patterns are typically E‑ and V‑heavy, reflecting long stretches of hypothesis generation and checking. For MATH and AIME25, conclusion-related patterns such as CC, CCC, and CCCC become substantially more prevalent, indicating that once a solution path is established, models often enter extended conclusion phases. In contrast, MBPP is strongly dominated by verification loops (VV, VVV, VVVV), suggesting repeated checking behavior in code-oriented reasoning.

Model size also systematically affects the distribution of reasoning patterns. Larger models (32B, 14B) exhibit a richer mix of transitions between behaviors (e.g., VE, EV, EVEE, VEVV, VVVC), whereas smaller models (8B, 4B) show higher frequencies of pure repetition such as EEEE and VVVV and of long homogeneous sequences like EEE and VVV. This suggests that larger models tend to interleave exploration, verification, and conclusion more fluidly, while smaller models rely more heavily on simple, repetitive behavior loops.

\begin{table*}[ht]
  \centering
  \small
  \setlength{\tabcolsep}{5pt}
  \caption{Frequencies of Top 20 3-gram patterns across models and datasets.}
  \label{tab:3gram}
  \begin{tabular}{l*{2}{cccc}}
    \toprule
    & \multicolumn{4}{c}{Qwen3-32B}
    & \multicolumn{4}{c}{Qwen3-14B} \\
    \cmidrule(lr){2-5}
    \cmidrule(lr){6-9}
    Rank
      & GPQA & MATH & AIME25 & MBPP
      & GPQA & MATH & AIME25 & MBPP \\
    \midrule
    \textbf{Top 1}
      & \cellcolor{Ecol!40}EEE (1.67) & \cellcolor{Vcol!40}VVV (1.04) & \cellcolor{Ccol!40}CCC (3.86) & \cellcolor{Vcol!40}VVV (1.07)
      & \cellcolor{Ecol!40}EEE (2.11) & \cellcolor{Vcol!40}VVV (1.58) & \cellcolor{Ecol!40}EEE (4.72) & \cellcolor{Vcol!40}VVV (1.42) \\
    \textbf{Top 2}
      & \cellcolor{Vcol!40}VVV (1.32) & \cellcolor{Ecol!40}EEE (0.89) & \cellcolor{Ecol!40}EEE (3.49) & \cellcolor{Ecol!40}EEE (0.53)
      & \cellcolor{Vcol!40}VVV (1.95) & \cellcolor{Ecol!40}EEE (1.21) & \cellcolor{Vcol!40}VVV (4.33) & \cellcolor{Ecol!40}EEE (0.81) \\
    \textbf{Top 3}
      & VEE (1.04) & \cellcolor{Ccol!40}CCC (0.85) & \cellcolor{Vcol!40}VVV (3.27) & PVV (0.50)
      & EVE (1.84) & \cellcolor{Pcol!40}PPP (0.85) & \cellcolor{Ccol!40}CCC (2.81) & VVE (0.52) \\
    \textbf{Top 4}
      & EVE (1.04) & \cellcolor{Pcol!40}PPP (0.80) & VCC (2.21) & VVC (0.43)
      & VEE (1.71) & VEE (0.80) & VVC (2.23) & VEV (0.50) \\
    \textbf{Top 5}
      & EEV (0.93) & VCC (0.62) & VVC (2.05) & OPV (0.42)
      & VEV (1.57) & VVE (0.71) & VCV (2.11) & VEE (0.49) \\
    \textbf{Top 6}
      & VVE (0.85) & VVC (0.55) & CVV (2.01) & VVE (0.41)
      & EEV (1.57) & VEV (0.60) & CVV (2.05) & PVV (0.49) \\
    \textbf{Top 7}
      & VEV (0.84) & ECC (0.54) & CCV (1.99) & VEE (0.37)
      & VVE (1.36) & EVE (0.60) & VCC (2.04) & EVE (0.44) \\
    \textbf{Top 8}
      & \cellcolor{Pcol!40}PPP (0.78) & CVC (0.53) & CVC (1.79) & VEV (0.37)
      & EVV (1.13) & VVC (0.60) & CVC (1.83) & VVC (0.42) \\
    \textbf{Top 9}
      & PVV (0.74) & OPP (0.53) & VCV (1.62) & EVE (0.30)
      & PVV (0.83) & EEV (0.59) & \cellcolor{Pcol!40}PPP (1.81) & EEV (0.40) \\
    \textbf{Top 10}
      & EVV (0.68) & VCV (0.50) & \cellcolor{Pcol!40}PPP (1.57) & EVV (0.29)
      & \cellcolor{Pcol!40}PPP (0.62) & CVV (0.58) & EEV (1.75) & EVV (0.39) \\
    \textbf{Top 11}
      & PPV (0.62) & VEE (0.50) & PCC (1.54) & VVP (0.29)
      & PPV (0.59) & VCV (0.57) & VEE (1.70) & OPV (0.38) \\
    \textbf{Top 12}
      & VPV (0.47) & CVV (0.49) & CCP (1.44) & \cellcolor{Pcol!40}PPP (0.28)
      & PVE (0.58) & \cellcolor{Ccol!40}CCC (0.57) & CCV (1.65) & OVV (0.24) \\
    \textbf{Top 13}
      & PEE (0.46) & PPC (0.49) & PPC (1.25) & EEV (0.28)
      & PEE (0.52) & PPV (0.55) & EVE (1.54) & VVP (0.24) \\
    \textbf{Top 14}
      & VVP (0.46) & PPV (0.48) & VEE (1.18) & VPV (0.27)
      & VPV (0.50) & PVV (0.51) & VVE (1.23) & VPV (0.23) \\
    \textbf{Top 15}
      & PVP (0.45) & PCC (0.46) & EEV (1.15) & OVV (0.26)
      & VVP (0.48) & VCC (0.50) & PPC (1.21) & PPV (0.15) \\
    \textbf{Top 16}
      & PVE (0.43) & CCV (0.45) & CPP (1.10) & PPV (0.26)
      & EEP (0.42) & OPP (0.50) & PCC (1.17) & PVE (0.13) \\
    \textbf{Top 17}
      & OPP (0.43) & VVE (0.42) & CPC (1.10) & VPP (0.23)
      & OPV (0.40) & EVV (0.48) & CCP (1.15) & VEC (0.13) \\
    \textbf{Top 18}
      & EEP (0.41) & EEC (0.41) & PCP (1.02) & PVP (0.22)
      & EPE (0.39) & ECC (0.48) & VEV (1.13) & VPP (0.13) \\
    \textbf{Top 19}
      & VPP (0.39) & PVV (0.40) & PPV (0.95) & PVE (0.19)
      & OPP (0.37) & EEC (0.47) & PVV (1.11) & PVP (0.12) \\
    \textbf{Top 20}
      & EPE (0.35) & VEC (0.39) & ECC (0.94) & VEC (0.17)
      & PVP (0.36) & PPC (0.44) & CPP (1.11) & EEC (0.12) \\
    \midrule
    & \multicolumn{4}{c}{Qwen3-8B}
    & \multicolumn{4}{c}{Qwen3-4B} \\
    \cmidrule(lr){2-5}
    \cmidrule(lr){6-9}
    \textbf{Top 1}
      & \cellcolor{Ecol!40}EEE (3.63) & \cellcolor{Vcol!40}VVV (1.94) & \cellcolor{Ecol!40}EEE (6.01) & \cellcolor{Vcol!40}VVV (1.98)
      & \cellcolor{Ecol!40}EEE (6.28) & \cellcolor{Ecol!40}EEE (2.61) & \cellcolor{Ecol!40}EEE (8.15) & \cellcolor{Vcol!40}VVV (1.52) \\
    \textbf{Top 2}
      & \cellcolor{Vcol!40}VVV (2.85) & \cellcolor{Ecol!40}EEE (1.78) & \cellcolor{Vcol!40}VVV (4.60) & \cellcolor{Ecol!40}EEE (0.95)
      & EVE (2.86) & \cellcolor{Vcol!40}VVV (1.48) & \cellcolor{Vcol!40}VVV (4.21) & \cellcolor{Ecol!40}EEE (1.40) \\
    \textbf{Top 3}
      & EVE (2.64) & VEE (1.07) & \cellcolor{Ccol!40}CCC (3.53) & VEV (0.68)
      & VEE (2.81) & VEE (1.19) & \cellcolor{Ccol!40}CCC (3.21) & VEE (0.65) \\
    \textbf{Top 4}
      & VEE (2.47) & \cellcolor{Pcol!40}PPP (0.88) & VVC (2.67) & VEE (0.66)
      & EEV (2.60) & EEV (0.89) & VVC (2.37) & VEO (0.62) \\
    \textbf{Top 5}
      & EEV (2.28) & VVE (0.88) & CVV (2.60) & VVE (0.64)
      & \cellcolor{Vcol!40}VVV (1.76) & EVE (0.88) & CVV (2.34) & VEV (0.58) \\
    \textbf{Top 6}
      & VEV (2.16) & VVE (0.88) & VCC (2.50) & EVE (0.63)
      & VEE (2.81) & VEE (1.19) & VEE (2.31) & VEE (0.65) \\
    \textbf{Top 7}
      & VVE (1.88) & \cellcolor{Ccol!40}CCC (0.80) & VCV (2.33) & PVV (0.57)
      & VEV (1.68) & EVE (0.88) & VCC (2.22) & EVE (0.62) \\
    \textbf{Top 8}
      & EVV (1.56) & EVE (0.79) & VEE (2.22) & EEV (0.52)
      & VVE (1.53) & VVE (0.84) & VCV (2.01) & VVE (0.57) \\
    \textbf{Top 9}
      & PVV (1.07) & EEV (0.78) & CCV (2.20) & VVC (0.50)
      & EVV (1.30) & EEC (0.75) & EEV (1.99) & EEV (0.54) \\
    \textbf{Top 10}
      & PVE (0.74) & CVV (0.75) & CVC (2.18) & EVV (0.49)
      & PEE (0.89) & \cellcolor{Pcol!40}PPP (0.72) & CCV (1.88) & PVV (0.52) \\
    \textbf{Top 11}
      & \cellcolor{Pcol!40}PPP (0.73) & VEV (0.74) & EEV (2.03) & OPV (0.45)
      & EEP (0.74) & VEV (0.70) & CVC (1.80) & OPV (0.48) \\
    \textbf{Top 12}
      & VPV (0.73) & VVC (0.73) & \cellcolor{Pcol!40}PPP (1.98) & OVV (0.26)
      & EPE (0.71) & \cellcolor{Ccol!40}CCC (0.69) & EVE (1.79) & EVV (0.46) \\
    \textbf{Top 13}
      & PEE (0.72) & VCV (0.73) & EVE (1.82) & VPV (0.25)
      & PVV (0.66) & ECC (0.62) & VVE (1.64) & OVV (0.22) \\
    \textbf{Top 14}
      & PPV (0.71) & EEC (0.64) & VVE (1.82) & VVP (0.25)
      & PVE (0.62) & CVV (0.61) & \cellcolor{Pcol!40}PPP (1.60) & VPV (0.21) \\
    \textbf{Top 15}
      & VVP (0.65) & VCC (0.63) & VEV (1.63) & PPV (0.18)
      & PPV (0.48) & VCV (0.60) & EEC (1.59) & VVP (0.20) \\
    \textbf{Top 16}
      & EEP (0.61) & ECC (0.63) & EVV (1.47) & PVE (0.18)
      & EEC (0.46) & VVC (0.60) & VEV (1.40) & VEC (0.17) \\
    \textbf{Top 17}
      & EPE (0.55) & EVV (0.60) & PPC (1.43) & EEC (0.17)
      & \cellcolor{Pcol!40}PPP (0.46) & EVV (0.57) & PPC (1.37) & PVE (0.17) \\
    \textbf{Top 18}
      & EVP (0.50) & CVE (0.58) & PCC (1.39) & VEC (0.16)
      & OPV (0.46) & CVE (0.54) & CCP (1.36) & EEC (0.17) \\
    \textbf{Top 19}
      & VPP (0.47) & PVV (0.57) & CCP (1.35) & EVC (0.15)
      & PEV (0.40) & PPV (0.49) & PCC (1.36) & EVC (0.13) \\
    \textbf{Top 20}
      & PVP (0.46) & CVC (0.57) & EEC (1.32) & PVP (0.14)
      & VEP (0.39) & PVV (0.48) & EVV (1.32) & PPV (0.12) \\
    \bottomrule
  \end{tabular}
\end{table*}

\begin{table*}[ht]
  \centering
  \small
  \setlength{\tabcolsep}{3pt}
  \caption{Frequencies of Top 20 4-gram patterns across models and datasets.}
  \label{tab:4gram}
  \begin{tabular}{l*{2}{cccc}}
    \toprule
    & \multicolumn{4}{c}{Qwen3-32B}
    & \multicolumn{4}{c}{Qwen3-14B} \\
    \cmidrule(lr){2-5}
    \cmidrule(lr){6-9}
    Rank
      & GPQA & MATH & AIME25 & MBPP
      & GPQA & MATH & AIME25 & MBPP \\
    \midrule
    \textbf{Top 1}
      & \cellcolor{Ecol!40}EEEE (0.93) & \cellcolor{Vcol!40}VVVV (0.53) & \cellcolor{Ecol!40}EEEE (2.06) & \cellcolor{Vcol!40}VVVV (0.54)
      & \cellcolor{Ecol!40}EEEE (1.07) & \cellcolor{Vcol!40}VVVV (0.87) & \cellcolor{Ecol!40}EEEE (2.81) & \cellcolor{Vcol!40}VVVV (0.80) \\
    \textbf{Top 2}
      & \cellcolor{Vcol!40}VVVV (0.63) & \cellcolor{Ecol!40}EEEE (0.45) & \cellcolor{Ccol!40}CCCC (1.75) & \cellcolor{Ecol!40}EEEE (0.31)
      & \cellcolor{Vcol!40}VVVV (0.99) & \cellcolor{Ecol!40}EEEE (0.63) & \cellcolor{Vcol!40}VVVV (2.32) & \cellcolor{Ecol!40}EEEE (0.51) \\
    \textbf{Top 3}
      & EVEE (0.52) & \cellcolor{Pcol!40}PPPP (0.32) & \cellcolor{Vcol!40}VVVV (1.65) & OPVV (0.25)
      & EVEE (0.88) & VEEE (0.36) & \cellcolor{Ccol!40}CCCC (1.15) & PVVV (0.27) \\
    \textbf{Top 4}
      & EEVE (0.49) & \cellcolor{Ccol!40}CCCC (0.31) & VCCC (0.98) & PVVV (0.25)
      & EEVE (0.88) & \cellcolor{Pcol!40}PPPP (0.34) & VVVC (1.14) & OPVV (0.27) \\
    \textbf{Top 5}
      & VEEE (0.48) & OPPP (0.25) & VVVC (0.97) & VVVC (0.21)
      & VEEE (0.78) & VVVE (0.30) & CVVV (0.99) & VEEE (0.26) \\
    \textbf{Top 6}
      & EEEV (0.42) & VVVC (0.24) & CVVV (0.91) & VVVE (0.18)
      & VEVE (0.78) & EVEE (0.28) & EEEV (0.90) & VVVC (0.25) \\
    \textbf{Top 7}
      & VEVE (0.39) & CVCC (0.23) & CCVV (0.83) & VEEE (0.17)
      & EEEV (0.72) & VVVC (0.28) & VCCC (0.90) & VVEV (0.24) \\
    \textbf{Top 8}
      & VEEV (0.37) & VEEE (0.23) & CVCC (0.79) & VVEV (0.15)
      & EVEV (0.71) & VVCV (0.27) & VEEE (0.87) & VVVE (0.24) \\
    \textbf{Top 9}
      & \cellcolor{Pcol!40}PPPP (0.33) & CVVV (0.21) & CCCV (0.77) & EVEE (0.15)
      & VEEV (0.65) & EEVE (0.27) & VVCV (0.85) & EVEE (0.24) \\
    \textbf{Top 10}
      & VVVE (0.33) & VVCV (0.20) & VVCC (0.76) & VEVV (0.15)
      & VVEV (0.60) & EEEV (0.26) & VCVV (0.83) & VEVV (0.21) \\
    \textbf{Top 11}
      & VVEV (0.33) & VCCC (0.20) & VVCV (0.72) & VVVP (0.14)
      & VVVE (0.55) & CVVV (0.26) & VVCC (0.82) & EEVE (0.21) \\
    \textbf{Top 12}
      & EVEV (0.33) & PPPC (0.20) & CCCP (0.72) & OVVV (0.14)
      & VEVV (0.55) & VVEV (0.26) & \cellcolor{Pcol!40}PPPP (0.81) & VEVE (0.20) \\
    \textbf{Top 13}
      & VVEE (0.32) & PPPV (0.20) & CVVC (0.70) & EEVE (0.14)
      & VVEE (0.54) & VVEE (0.25) & EEVE (0.81) & EEEV (0.20) \\
    \textbf{Top 14}
      & PVVV (0.32) & VVCC (0.19) & CCVC (0.70) & VVEE (0.13)
      & EVVE (0.53) & OPPP (0.24) & EVEE (0.75) & VVEE (0.19) \\
    \textbf{Top 15}
      & VEVV (0.30) & EEEV (0.18) & PCCC (0.70) & PPPP (0.13)
      & EEVV (0.46) & VCVV (0.24) & VCVC (0.72) & EVVV (0.19) \\
    \textbf{Top 16}
      & EVVE (0.29) & EECC (0.18) & VCCV (0.65) & EEEV (0.12)
      & EVVV (0.42) & VEVV (0.22) & CVCV (0.71) & VEEV (0.17) \\
    \textbf{Top 17}
      & EEVV (0.27) & VCVC (0.18) & VCVC (0.61) & EVVV (0.12)
      & PVVV (0.40) & VEVE (0.22) & CCCV (0.70) & EVEV (0.15) \\
    \textbf{Top 18}
      & EVVV (0.26) & VVVE (0.17) & VCVV (0.61) & VEVE (0.12)
      & \cellcolor{Pcol!40}PPPP (0.28) & PPPV (0.22) & CVCC (0.69) & OVVV (0.14) \\
    \textbf{Top 19}
      & PPVV (0.25) & EEEC (0.17) & EEEV (0.60) & VPVV (0.12)
      & PVEE (0.25) & PVVV (0.22) & CCVC (0.67) & EVVE (0.13) \\
    \textbf{Top 20}
      & PPPV (0.24) & VCVV (0.17) & \cellcolor{Pcol!40}PPPP (0.60) & VEEV (0.12)
      & PPVV (0.25) & EVVV (0.21) & CVVC (0.61) & VVVP (0.13) \\
    \midrule
    & \multicolumn{4}{c}{Qwen3-8B}
    & \multicolumn{4}{c}{Qwen3-4B} \\
    \cmidrule(lr){2-5}
    \cmidrule(lr){6-9}
    \textbf{Top 1}
      & \cellcolor{Ecol!40}EEEE (2.04) & \cellcolor{Vcol!40}VVVV (1.09) & \cellcolor{Ecol!40}EEEE (3.51) & \cellcolor{Vcol!40}VVVV (1.21)
      & \cellcolor{Ecol!40}EEEE (4.05) & \cellcolor{Ecol!40}EEEE (1.58) & \cellcolor{Ecol!40}EEEE (5.21) & \cellcolor{Ecol!40}EEEE (0.98) \\
    \textbf{Top 2}
      & \cellcolor{Vcol!40}VVVV (1.56) & \cellcolor{Ecol!40}EEEE (0.98) & \cellcolor{Vcol!40}VVVV (2.23) & \cellcolor{Ecol!40}EEEE (0.59)
      & EEVE (1.69) & \cellcolor{Vcol!40}VVVV (0.79) & \cellcolor{Vcol!40}VVVV (2.13) & \cellcolor{Vcol!40}VVVV (0.86) \\
    \textbf{Top 3}
      & EEVE (1.36) & VEEE (0.48) & \cellcolor{Ccol!40}CCCC (1.63) & PVVV (0.35)
      & EVEE (1.68) & VEEE (0.61) & \cellcolor{Ccol!40}CCCC (1.36) & VEEE (0.36) \\
    \textbf{Top 4}
      & EVEE (1.33) & EVEE (0.37) & VVVC (1.26) & EVEE (0.33)
      & VEEE (1.63) & EEEV (0.47) & VEEE (1.26) & EVEE (0.35) \\
    \textbf{Top 5}
      & VEEE (1.18) & \cellcolor{Pcol!40}PPPP (0.36) & VEEE (1.21) & VVVE (0.32)
      & EEEV (1.51) & EVEE (0.47) & EEEV (1.21) & EEVE (0.32) \\
    \textbf{Top 6}
      & EEEV (1.07) & EEEV (0.36) & CVVV (1.19) & VEEE (0.32)
      & VEVE (0.91) & EEVE (0.45) & VVVC (1.07) & OPVV (0.31) \\
    \textbf{Top 7}
      & VEVE (1.05) & VVVE (0.35) & EEEV (1.11) & OPVV (0.31)
      & \cellcolor{Vcol!40}VVVV (0.88) & EEEC (0.38) & CVVV (1.04) & PVVV (0.29) \\
    \textbf{Top 8}
      & VEEV (0.98) & EEVE (0.35) & VVCV (1.06) & EEVE (0.30)
      & VEEV (0.85) & VVEE (0.35) & EEVE (1.01) & VVVC (0.29) \\
    \textbf{Top 9}
      & EVEV (0.98) & VVVC (0.34) & VCCC (1.03) & VVVC (0.30)
      & EVEV (0.85) & VVVE (0.34) & EVEE (0.97) & VVEV (0.25) \\
    \textbf{Top 10}
      & VVEV (0.84) & VVEE (0.34) & EEVE (1.01) & VVEV (0.29)
      & VVEE (0.74) & VEVE (0.29) & VVCV (0.97) & VEVE (0.25) \\
    \textbf{Top 11}
      & VEVV (0.78) & VVCV (0.32) & CVCC (0.99) & VEVE (0.29)
      & EVVE (0.65) & VVEV (0.29) & EEEC (0.93) & VEVV (0.25) \\
    \textbf{Top 12}
      & VVVE (0.77) & CVVV (0.31) & VVCC (0.97) & VEVV (0.29)
      & EEVV (0.61) & VEEV (0.28) & VCVV (0.89) & VEVV (0.24) \\
    \textbf{Top 13}
      & VVEE (0.74) & VCVV (0.30) & VCVV (0.97) & VEVV (0.29)
      & VVVE (0.58) & EECC (0.28) & VVCC (0.87) & VEEV (0.22) \\
    \textbf{Top 14}
      & EVVE (0.71) & VVEV (0.30) & CCVV (0.94) & VEEV (0.25)
      & VVEV (0.58) & \cellcolor{Pcol!40}PPPP (0.28) & PEEE (0.86) & EVVV (0.21) \\
    \textbf{Top 15}
      & EVVV (0.60) & EEEC (0.30) & EVEE (0.93) & EVVV (0.24)
      & VEVV (0.55) & VVCV (0.27) & VCCC (0.84) & VVEE (0.20) \\
    \textbf{Top 16}
      & EEVV (0.60) & VEEV (0.30) & VCVC (0.89) & VVEE (0.23)
      & PEEE (0.54) & CVVV (0.27) & EEEP (0.80) & EVEV (0.20) \\
    \textbf{Top 17}
      & PVVV (0.53) & VEVE (0.29) & \cellcolor{Pcol!40}PPPP (0.81) & EVEV (0.23)
      & EVVV (0.46) & VVVC (0.26) & CVVC (0.79) & EVVE (0.16) \\
    \textbf{Top 18}
      & PEEE (0.39) & \cellcolor{Ccol!40}CCCC (0.28) & VCCV (0.81) & EVVE (0.16)
      & EPEE (0.43) & \cellcolor{Ccol!40}CCCC (0.25) & CEEE (0.79) & EEVV (0.15) \\
    \textbf{Top 19}
      & PVEE (0.35) & VEVV (0.27) & VVVE (0.81) & OVVV (0.15)
      & EEEP (0.43) & CVEE (0.25) & CVCC (0.77) & OVVV (0.13) \\
    \textbf{Top 20}
      & VPVV (0.34) & PVVV (0.27) & CCCV (0.79) & VVVP (0.14)
      & EEPE (0.43) & VCVV (0.24) & CCVV (0.75) & OVVV (0.13) \\
    \bottomrule
  \end{tabular}
\end{table*}

\end{document}